\documentclass[sigconf]{acmart}

\usepackage{booktabs} %
\def \sys {\textit{DeepLoc}}
\usepackage{subcaption}
\usepackage{pbox}
\usepackage{amsmath}
\usepackage{flushend}
\setcopyright{rightsretained}

\acmDOI{10.475/123_4}

\acmISBN{123-4567-24-567/08/06}

\acmConference[SIGSPATIAL'18]{ACM Sigspatial conference}{June 2018}{Seattle, Washington, USA}
\acmYear{2018}
\copyrightyear{2018}

\acmArticle{4}
\acmPrice{15.00}
\usepackage[outdir=Figures/]{epstopdf}
\graphicspath{{Figures/}}
\begin{document}
\title{DeepLoc: A Ubiquitous Accurate and Low-Overhead Outdoor Cellular Localization System}

\author{Ahmed Shokry}
\orcid{1234-5678-9012}
\affiliation{%
  \institution{Alexandria University}
  \streetaddress{}
  \city{Alexandria}
  \country{Egypt}
  \state{}
  \postcode{}
}
\email{ahmed.shokry@alexu.edu.eg}

\author{Marwan Torki}
\affiliation{%
  \institution{Alexandria University}
  \streetaddress{P.O. Box 1212}
  \city{Alexandria}
  \country{Egypt}
  \state{}
  \postcode{}
}
\email{mtorki@alexu.edu.eg}

\author{Moustafa Youssef}
\affiliation{%
  \institution{Egypt-Japan Univ. of Sc. \& Tech.}
  \streetaddress{}
  \city{Alexandria}
  \country{Egypt}}
\email{moustafa.youssef@ejust.com}

\renewcommand{\shortauthors}{Ahmed Shokry, Marwan Torki, Moustafa Youssef.}
\def \sys {\textit{DeepLoc}}

\begin{abstract}

Recent years have witnessed fast growth in outdoor location-based services. While GPS is considered a ubiquitous localization system, it is not supported by low-end phones, requires direct line of sight to the satellites, and can drain the phone battery quickly.

In this paper, we propose \sys{}: a deep learning-based outdoor localization system that obtains GPS-like localization accuracy without its limitations. In particular, \sys{} leverages the ubiquitous cellular signals received from the different cell towers heard by the mobile device as hints to localize it. To do that, crowd-sensed geo-tagged received signal strength information coming from different cell towers is used to train a deep model that is used to infer the user's position. As part of \sys{} design, we introduce modules to address a number of practical challenges including scaling the data collection to large areas, handling the inherent noise in the cellular signal and geo-tagged data, as well as providing enough data that is required for deep learning models with low-overhead. 

We implemented \sys{} on different Android devices. Evaluation results in realistic urban and rural environments show that \sys{} can achieve a median localization accuracy within 18.8m in urban areas and within 15.7m in rural areas. 
This accuracy outperforms the state-of-the-art cellular-based systems by more than 470\% and comes with 330\% savings in power compared to the GPS. This highlights the promise of \sys{} as a ubiquitous accurate and low-overhead localization system. 
\end{abstract}



\copyrightyear{2018} 
\acmYear{2018} 
\setcopyright{acmcopyright}
\acmConference[SIGSPATIAL '18]{26th ACM SIGSPATIAL International Conference on Advances in Geographic Information Systems}{November 6--9, 2018}{Seattle, WA, USA}
\acmBooktitle{26th ACM SIGSPATIAL International Conference on Advances in Geographic Information Systems (SIGSPATIAL '18), November 6--9, 2018, Seattle, WA, USA}
\acmPrice{15.00}
\acmDOI{10.1145/3274895.3274909}
\acmISBN{978-1-4503-5889-7/18/11}

\maketitle

\section{Introduction}
Outdoor localization systems have gained focus recently in different application domains including navigation systems, location-based advertisements, and location-based social networks \cite{elhamshary2014checkinside,shankar2012crowds,zheng2011location}. Generally, GPS \cite{hofmann2012global} is considered the most commonly used system due to its availability worldwide. However, GPS requires a direct line-of-sight to the satellites, which limits its accuracy and availability in urban areas and bad weather conditions outdoors \cite{cui2003autonomous}. In addition, GPS is power-hungry, which can drain the battery of the energy-limited mobile devices quickly \cite{gaonkar2008micro}. To mitigate the high-energy requirement of the GPS, several techniques are proposed \cite{youssef2010gac,constandache2010compacc,jurdak2010adaptive,paek2010energy} that use mobile sensors with duty-cycling of the GPS to trade-off localization accuracy with energy-efficiency. To address the shortcomings of GPS-based techniques, WiFi-based outdoor localization systems have been proposed, e.g. \cite{cheng2005accuracy,lamarca2005place,Skyhook,thiagarajan2009vtrack}. These systems can estimate the user's position using the ambient WiFi signals overheard from access points deployed in buildings along the roads. However, WiFi signals are not available in many road areas, e.g. on highways and rural areas. In addition, WiFi signals outside buildings are usually weak, limiting their coverage and ability to differentiate distinct locations.

\par
To further boost accuracy and reduce the energy consumption, a number of systems have been proposed recently that leverage smart phones augmented sensors with low-energy profile, such as the compass, gyroscope, and accelerometer, for localization  \cite{aly2013dejavu,aly2017dejavuj,aly2015lanequest,aly2016lanequestj,constandache2010towards,ofstad2008aampl,azizyan2009surroundsense,arthi2010localization,light}. These systems can provide high-accuracy low-energy outdoor localization. \textit{However, similar to GPS and WiFi-based systems, these systems are not available in low-end phones that are still used throughout the world, limiting their ubiquitousness.}
\par

To provide a real ubiquitous localization service, a number of systems have been introduced that are based on cellular signals, which are available by default in any cell phone with zero extra energy in addition to the standard phone operation \cite{ibrahim2012cellsense,ibrahim2011hidden,elnahrawy2007adding,li2008angle,varshavsky2005gsm}. These techniques either depend on propagation models or fingerprinting for estimating the user location 
\cite{vo2016survey}. 
The propagation models-based techniques, e.g. \cite{elnahrawy2007adding,li2008angle,elbakly2016robust,elbakly2015calibration}, capture the relation between signal strength and distance.  
Due to signal reflection from buildings, interference, attenuation, and other factors that affect the signal propagation, propagation model-based techniques achieve limited localization accuracy outdoors
\cite{weiss2003accuracy,ghaboosi2011geometry}.
On the other hand, fingerprinting techniques, e.g. \cite{ibrahim2012cellsense,ibrahim2011hidden,ibrahim2010cellsense,ergen2014rssi}, usually work in two phases. The first phase is an offline phase to collect measurements such as the received signal strength (RSS) from the different cell towers at known locations. These measurements are then stored as signatures, i.e. fingerprint, for the different locations in the area of interest. The second phase is an online tracking phase to collect online measurements from a user at an unknown location and to estimate the user location as the best location in the fingerprint that matches the online measurements. Current fingerprinting techniques use graphical models like Bayesian networks \cite{ibrahim2012cellsense,ibrahim2010cellsense,ibrahim2010cellsense,paek2011energy} to estimate the user's position. The graphical models-based solutions try to learn the distribution of the received signal strength coming from each cell tower. To make the problem mathematically tractable and avoid the curse of dimensionality \cite{bishop2006pattern}, these systems usually assume that the different cell towers are \textit{independent}, limiting their ability to capture the inherently correlated relation between the RSS from the different cell towers, which in turn lowers their accuracy. 
\par
\par
In this paper, we propose \sys{}: a deep learning-based outdoor cellular localization system that can achieve GPS-like accuracy without its limitations. Specifically, \sys{} builds on deep learning to automatically capture the unique signatures of the different cell towers at different fingerprint locations, without assuming cell towers independence. It works in two phases: during the offline phase (model training phase), location-tagged cell towers RSS measurements are collected in a crowd-sensing manner by war-drivers or users with GPS-enabled phones\footnote{Note that GPS-enabled phones are only needed during the offline phase. The online phase is completely based on cellular information.}. These measurements are used to train a deep neural network model. In the online phase (tracking phase), a user standing at an unknown location scans the RSS coming from different cell towers (online measurements). These measurements are then fed into the deep neural network model to estimate the user's position.
\par
To achieve its goals, \sys{} needs to address a number challenges: First, traditional fingerprints, e.g. \cite{youssef2005horus,youssef2006location,youssef2003wlan,bahl2000radar}, are collected while the user remains still in different fingerprint locations for sometime, typically a few minutes. This does not scale to large outdoor areas. Second,  due to the complex propagation environment, the RSS samples are noisy and are not consistent. Third, similarly, reported GPS locations used for training the deep learning model during the offline phase have inherent errors that affect the fingerprint accuracy. Finally, deep learning models require a large amount of training data that may be a high overhead process. \sys{} design includes a number of innovative modules to mitigate these challenges including a gridding approach to trade-off accuracy and overhead, different data augmentation techniques to handle the noisy RSS and GPS data and reduce the deep learning data requirements, and model regularization to further reduce the noise effect and increase the system robustness.

We have implemented \sys{} on different Android phones and evaluated it in typical urban and rural areas. Our results show that \sys{} can achieve a consistent median localization error of 18.8m in urban areas and 15.7m in rural areas which are significantly better than the state-of-the-art cellular localization techniques by more than 470\% and 1300\% in urban and rural areas respectively. This comes with zero extra energy consumption compared to the normal phone operation and is more power efficient than GPS by 330\%. %
\par
The rest of the paper is organized as follows: Section~\ref{overview}
presents our system architecture. Section~\ref{details} gives the details
of the \sys{} system followed by its
evaluation in Section~\ref{evaluation}. Section~\ref{history} discusses related
work. Finally, we conclude the paper in Section~\ref{conclusion}.
\section{System Overview}
\label{overview}

\begin{figure}[t!]
	\centering
	\includegraphics[width=1.1\columnwidth]{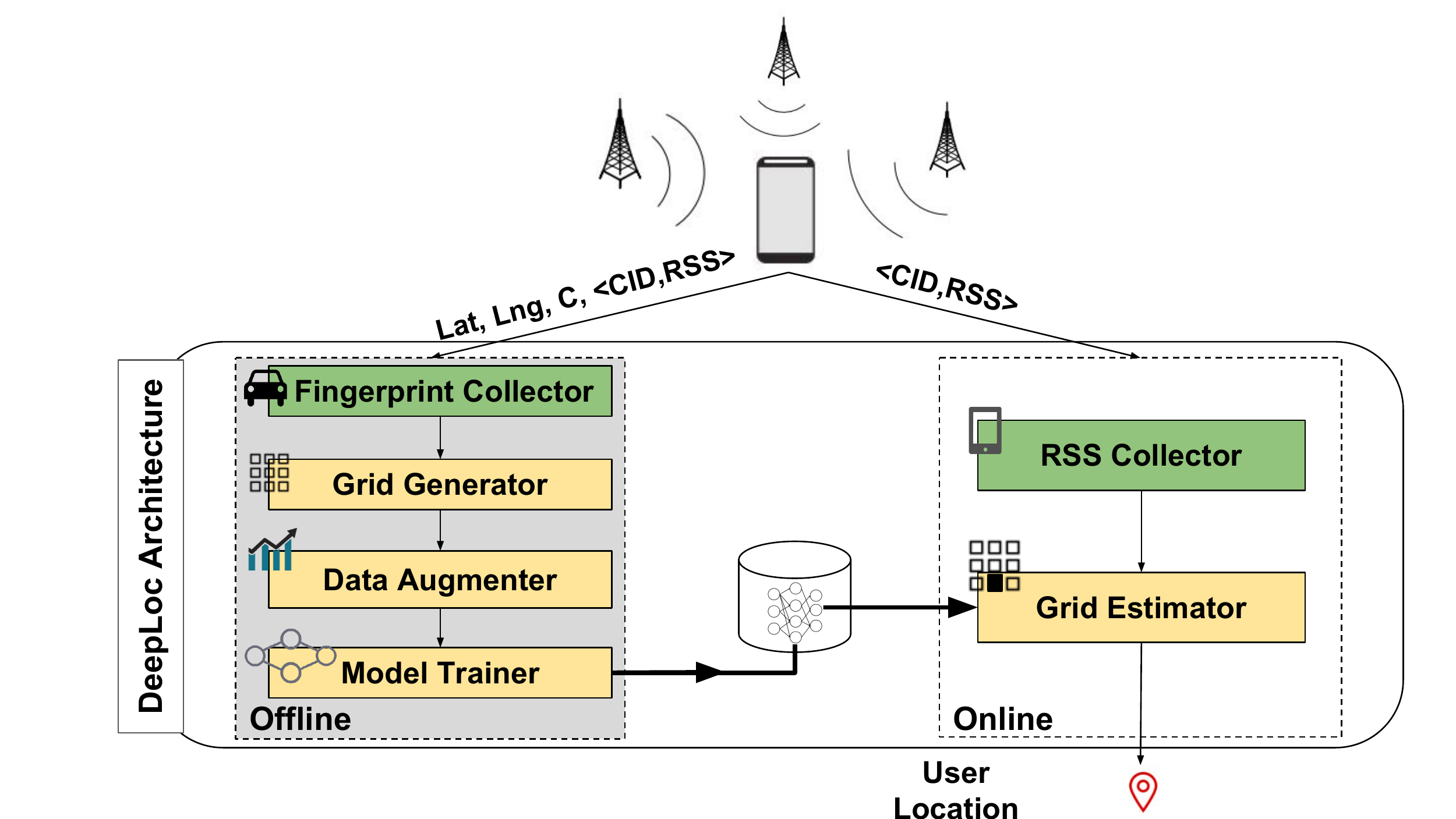}
	\caption{DeepLoc system architecture.}
	\label{arcitecture}
\end{figure}

In this section, we describe \sys{} main components and give an overview about its operation. Figure~\ref{arcitecture} shows the system architecture. \sys{} works in two phases: offline training phase and online phase. During the offline phase, \sys{} collects the RSS measurements through crowd-sensing from war-drivers or system users through the Fingerprint Collector module. The module is deployed on smart phones, including those with GPS, to collect the GPS location information along with the received signal strength (RSS) information simultaneously. To do that, it scans the RSS information from up to seven neighboring cell towers, including the associated cell tower. The collected RSS information is the cell towers IDs and their RSS measurements. The collected GPS location information is the latitude, longitude, and the confidence in the location coordinates, as reported by the operating system API. These measurements are then forwarded to the Grid Generator module. 
\par
Traditionally \cite{youssef2005horus,bahl2000radar}, to build the fingerprint, the data collectors have to stop and stay at the different fingerprint locations for enough time, typically a few minutes, to gather the RSS samples. This increases the fingerprint construction overhead \cite{shokry2017tale}, especially in large areas, which is the norm outdoors. To reduce this overhead, the Grid Generator module super-imposes a virtual grid over the area of interest. Data collectors move naturally during their daily life and the geo-tagged RSS samples collected inside each cell are used as training samples to build the deep model for this cell. 
\par
The Data Augmenter module is used to handle the noise in the training data including both in the input GPS locations as well as the noisy cellular RSS. It also increases the number of training samples, leading to both reduced data collection overhead and more robustness. %
\par
The Model Trainer module trains a deep neural network model using the RSS measurements that are associated with the grid information. The input to the model is the RSS from the different cell towers heard in the area of interest. The output is a probability distribution over the different grid cells in the area of interest. Special provisions are used to avoid over-training as well as to increase the model robustness. Note that the model is incrementally and dynamically updated as more training samples become available from the system users, allowing \sys{} to maintain a fresh fingerprint all the time.
\par
During the online tracking phase, a user standing at an unknown locations scans the RSS using the RSS Collector. This module collects the cell towers IDs and the received signal strength measurements and forwards them to the Location Estimator module. The Location Estimator module uses the trained deep network to calculate the probability distribution of each grid cell and fuses this information to estimate the user location.

\begin{table}[!t]
	\centering
	\caption{ Table of notations}
	\label{tab:notations}
	\scalebox{0.8}{
		
		\begin{tabular}{|c|l|}\hline
			Symbol & Description \\ \hline \hline
			$\mathbb{G}$ & Universe of grid cells in the area of interest (i.e virtual grid).\\ \hline
			$K$ & Number of grid cells. \\ \hline
			$M$ & Total number of cell towers in the environment. \\ \hline
			$N_s$ & Total number of training samples. \\ \hline
			$x_i$ & Model input (i.e training sample $|x_i|$ = $M$) where $1 \leq i \leq N_s$. \\ \hline
			$x_{ij}$ & Input RSS from cell tower $j$ at training sample $i$ where $1 \leq j \leq M$. \\ \hline
			$y_i$ & Model output (i.e logits $|y_i|$ = $K$) for corresponding input $x_i$.  \\ \hline
			$y_{ij}$ & Output score that scan $x_i$ is at grid cell $j$. \\ \hline
			$p(y_{ij})$ & Probability that scan $x_i$ is at grid cell $j$.\\ \hline
			$P(y_i)$ & Probability distribution for different grid cells ($|P|$ = $K$)).\\ \hline
			$L_i$ & \pbox{9cm}{One-hot encoded vector present the probability distribution for different grid cells in the training phase ($|L_i|$ = $K$)).} \\ \hline
			$x$ & Online RSS sample/scan vector $|x|=M$.\\ \hline
			$P(g|x)$ & Probability of receiving a signal vector $x$ at grid cell $g \in \mathbb{G}$.\\ \hline
			$g^*$ & Estimated grid cell.\\ \hline
			$l^*$ & Center of mass of $g^*$.\\ \hline
			\hline
			
		\end{tabular}
	}
\end{table}

\begin{figure}[t!]
	\centering
	\includegraphics[width=0.8\columnwidth]{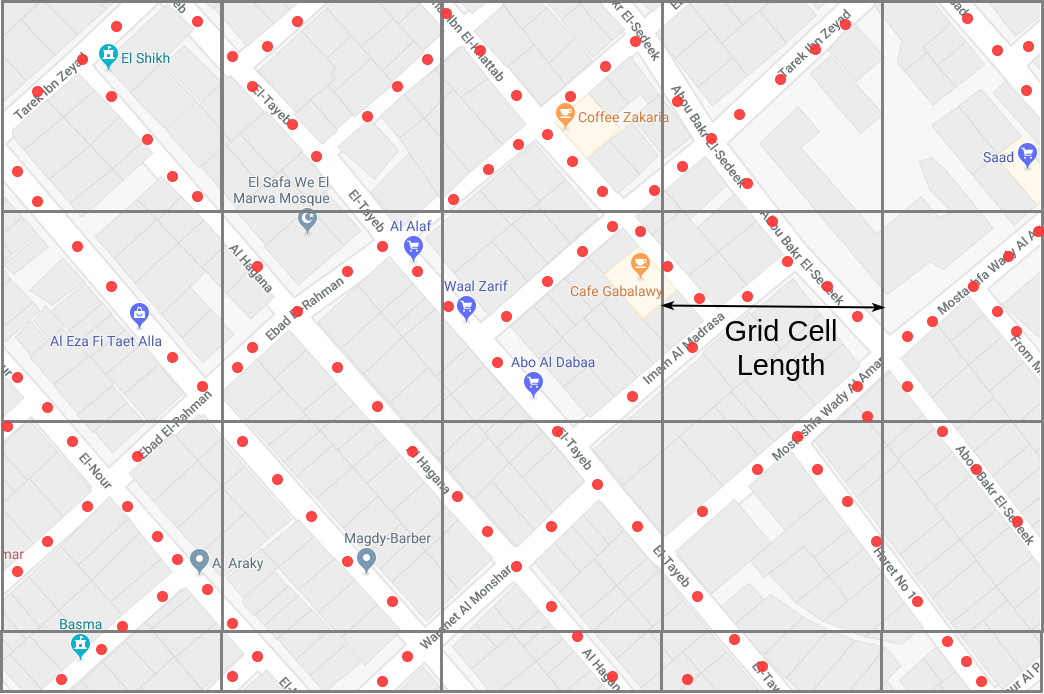}
	\caption{The gridding approach. The area of interest is superimposed by equally-sized square cells. Red points present the RSS samples.}
	\label{fig:grid} 
\end{figure}

\begin{figure*}[t!] 
	\begin{subfigure}[t]{0.33\linewidth}
		\centering
		\includegraphics[width=0.9\linewidth]{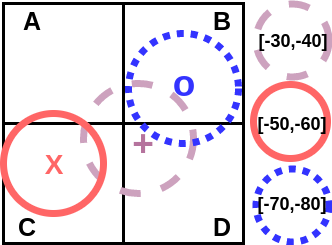} 
		\caption{Three RSS samples.}
		\label{fig:structureConf} 
	\end{subfigure}%
	\begin{subfigure}[t]{0.33\linewidth}
		\centering
		\includegraphics[width=0.9\linewidth]{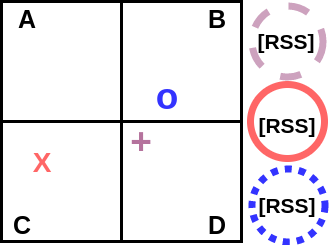} 
		\caption{Without augmentation.}
		\label{fig:aug-loconly} 
	\end{subfigure}%
	\begin{subfigure}[t]{0.33\linewidth}
		\centering
		\includegraphics[width=0.9\linewidth]{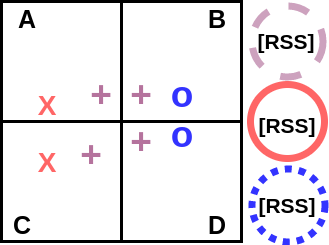} 
		\caption{Spatial augmentation.}
		\label{fig:aug-locconf} 
	\end{subfigure} 
	\caption{Illustration of the spatial augmentation technique. (a) Example showing ground-truth GPS estimated locations (centers of circles) for three RSS samples, the estimated confidence (circles), and the RSS
		vector from the two cell towers heard in each scan (the
		numbers inside the circles). The confidence circles intersect
		with the four grid cells A, B, C and D. (b) without augmentation, each sample is assigned to the grid cell its center falls in. (c) with spatial augmentation, each sample is assigned to \textit{all grid cells the confidence circle interests with}.}
	\label{fig:spatial_aug} 
\end{figure*}
\section{The DeepLoc System}
\label{details}
In this section, we present the \sys{} system details including the offline model training phase and the online user tracking phase. We also describe how the system handles different practical challenges. Table~\ref{tab:notations} summarizes the notations used in this section.
\subsection{The Offline Model Training Phase}
During this phase, the deep network model is trained taking into account a number of practical challenges including: how to reduce the data collection overhead, how to handle the noise in training data, and how to train the deep model while avoiding over-training and increasing the model robustness. 
\subsubsection{The grid generator} 
To remove the requirement of the data collectors standing still at different fingerprint locations for a certain time, the Grid Generator module divides the area of interest into a virtual grid (Figure~\ref{fig:grid}). Each training sample in the form of \textit{(latitude, longitude, confidence, cell-towers IDs, RSS information)} is mapped to a specific cell based on its coordinates. All the samples within a virtual grid cell are used as training samples for this cell signature. The cell length is a parameter that presents a trade-off between the localization accuracy and the computational-complexity as we quantify in Section~\ref{evaluation}.
\par
Note that this gridding approach tackles multiple scalability challenges. First, it allows the data to be collected while  the war-drivers or users are moving naturally in their life, without requiring them to stand still at the different fingerprint locations. Second, it provides a way to reduce the number of grid cells in the area of interest, and hence controls the model complexity/accuracy. 
\subsubsection{The data augmenter}
\label{sec:data_aug}

\begin{figure}[!t]
	\includegraphics[width=0.9\linewidth]{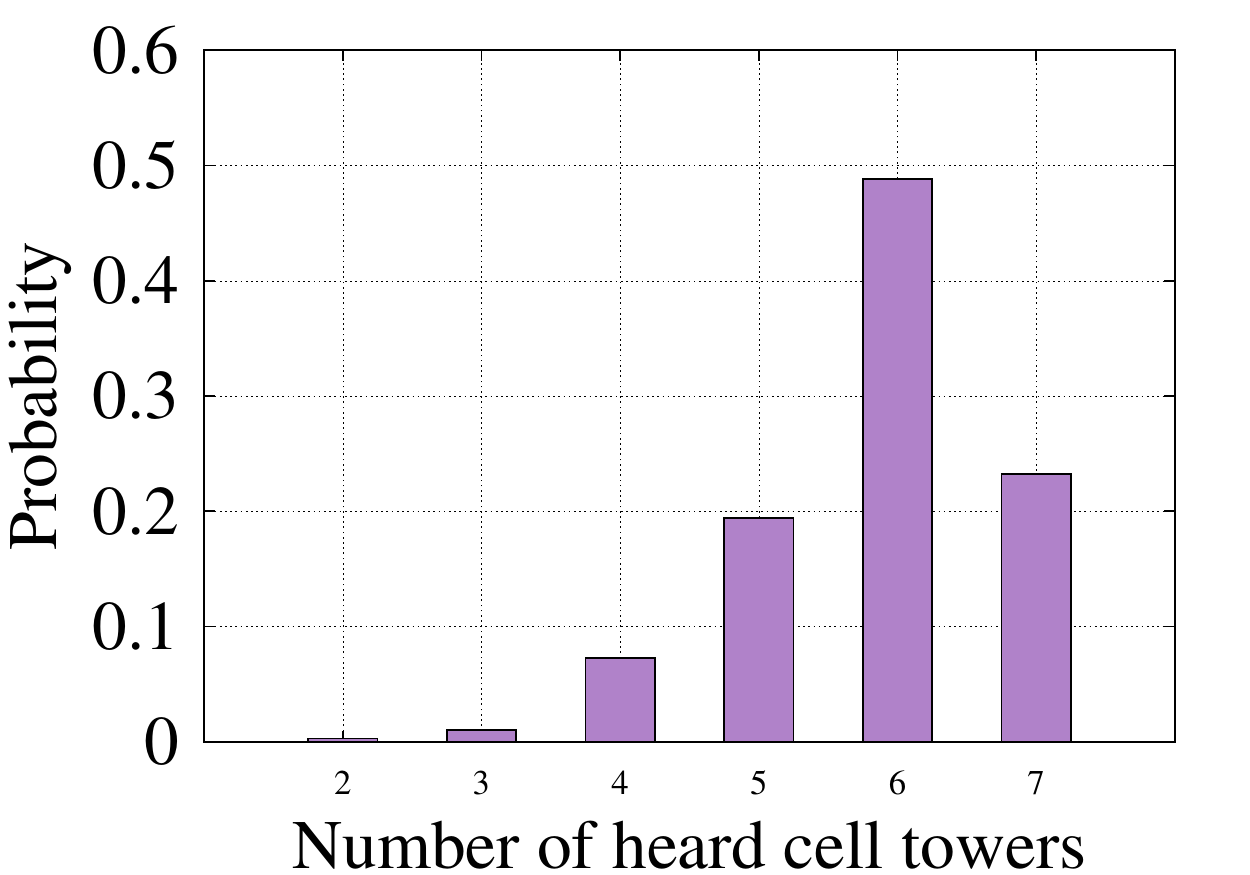}
	\caption{Probability of hearing different number of cell towers at one location.}%
	\label{fig:hist}
\end{figure}

\begin{figure}[t!] 
	\centering
	\includegraphics[width=1.0\linewidth]{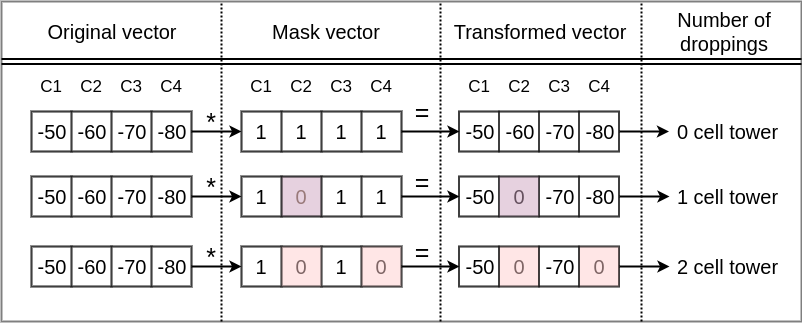} 
	\caption{Scans augmentation technique. The original scan vector is multiplied by a mask to obtain the transformed vector after dropping a number of cell towers RSS (setting their values to zero), emulating the noise in the input signal.}
	\label{fig:scan_aug} 
\end{figure}

In practice, the quality and the size of the training data influence the deep learning classification accuracy. The goal of the Data Augmenter module is to both handle the noise in the input data and automatically generate new training samples to increase the input data size, reduce the data collection overhead, and increase the system robustness.

\par
\sys{} has two data augmenters sub-modules: Spatial Augmenter to handle the noise in the GPS information and Scans Augmenter to handle the noise in the RSS data. Both modules increase the training set size as explained next. 

\noindent \textbf{The Spatial Data Augmenter:}
\par
GPS usually has inherent errors that can range from a few meters up to tens of meters in some cases \cite{aly2013dejavu,aly2017dejavuj}. This can lead to placing the collected RSS sample in an incorrect cell, affecting the system accuracy. To address this, \sys{} leverages the GPS confidence circle reported by the operating system API. The intuition is that since the reported GPS location is not accurate, the user can be actually located anywhere inside the confidence circle. Therefore, the \sys{} spatial augmentation module places the collected sample in all grid cells that intersect with the GPS confidence circle. This is particularity useful for samples collected near the cell boundaries. Figure~\ref{fig:spatial_aug} shows an example. 

Note that this also implicitly increases the number of training samples as each raw sample can be placed in multiple grid cells. 

\noindent \textbf{The Scan Data Augmenter:}
\par
The number of cell towers heard at a fixed location varies with time due to the noisy wireless channel. Figure~\ref{fig:hist} shows the probability mass function of the number of cell towers at a given location using our collected data. The figure shows that we can hear seven cell towers (maximum number of cell towers) with around 20\% probability only. 

To reduce the required number of samples per cell and to reflect the typical wireless channel characteristics with a small number of samples, \sys{} introduces the scan augmentation technique. The intuition here is to mimic the varying wireless channel by simulating the case of not hearing some cell towers at a given location in the different samples. This is only applied to scans that have more than five heard cell towers, where all the combinations of removing up to two cell towers are applied to generate new samples from the original full raw sample (Figure~\ref{fig:scan_aug}).

\par
Note that the spatial and scan augmentation techniques are independent and can be applied together. Therefore, we have a total of four different variations for data augmentation, corresponding to whether each augmentation technique is used or not. We quantify the effect of these different variations in Section~\ref{evaluation}.

Data augmentation helps in both handling the noise in the input data as well as providing more training samples and patterns to the deep network. This leads to better accuracy and robustness.

\subsubsection{The Model trainer}
\parbox{1cm}{  }
\par
\noindent \textbf{Basic Model}

\begin{figure}[!t]
	\includegraphics[width=0.9\linewidth]{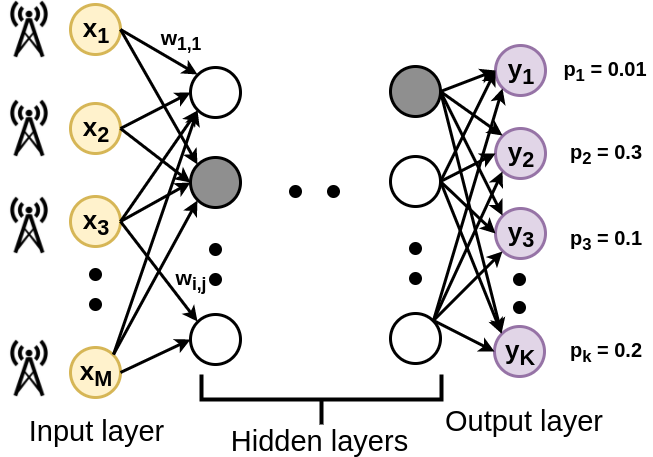}
	\caption{Network structure. The input is the RSS coming from different cell towers in the environment (features). The output is the probability distribution for different grid cells in the area of interest. Grey-shaded neurons represent examples of nodes that have been temporary dropped-off to increase the model robustness and avoid over-training.}
	\label{fig:structure}
\end{figure}

The target of this module is to train the deep neural network model using the \textit{augmented} fingerprint samples dataset constructed by the \textit{Data Augmenter} module. Figure~\ref{fig:structure} presents our deep network model structure. The input to the classifier is the RSS coming from $M$ cell-towers that cover the area of interest. The output is the grid cell probability distribution, i.e. the probability that the input sample/scan belongs to each cell in the area of interest.  

Each training sample consists of RSS information from different cell towers that cover the entire area of interest. 
If a cell tower is not heard in a scan, \sys{} sets its corresponding RSS to zero.
\sys{} employs a multinomial logistic model~\cite{hosmer2013applied}. 
This deep model acts as a non-linear function whose independent variables (i.e network input) is a set of real-valued received signal strengths coming from different cell towers deployed in the area of interest. The dependent variables (i.e network output) is a probability distribution over the virtual grid cells.
Since each input scan in the augmented data belongs to only one grid cell, the output probability distribution is presented as one-hot encoded vector with one at the grid cell the input training sample is in and zero otherwise.
The model can infer the probability distribution of different grid cells through learning the feature vector of the different training examples. 
\par
More formally, assume that we have $N_s$ training samples present our augmented fingerprint training set collected through corwd-sensing. 
Each record $x_i$ where $ 1\leq i \leq N_s$ contains $M$ features $(x_{i1},x_{i2},..x_{iM})$ that present the RSS information coming from $M$ cell towers covering the entire area of interest. The corresponding discrete outcomes (i.e logits) for the input $x_i$ is $y_i = (y_{i1},y_{i2},..,y_{iK})$ present the score for each grid cell of the possible $K$ grid cells to be the estimated cell. \sys{} employs the Softmax function to convert the logit score $y_{ij}$ (for sample $i$ to be at grid cell $j$) to a probability. The Softmax function calculates the score probability as follows
\begin{equation}
p(y_{ij}) = \frac{e^{y_{ij}}}{\sum_{j=1}^{K}e^{y_{ij}}}
\end{equation}
where 
\begin{equation}
\sum_{j=1}^{K} p(y_{ij}) = 1
\end{equation}
During the training phase, the expected/ground-truth probability vector $P(y_i) = [ p(y_{i1}) , p(y_{i2}) ...   p(y_{iK})]$ is presented as one-hot vector ($L_i$) which has one for the entry corresponding to the correct grid cell and zeros for other grid cells. 
To train the model we employ a stochastic gradient descent optimizer that attempts to minimize the average cross-entropy distance between the estimated output vector probability $P(y_i)$ and the one-hot vector $L_i$ over the entire training dataset. 
The loss function is defined as:
\begin{equation}
\mathrm{loss} = \frac{1}{N_s} \sum_{i=1}^{N_s} D(P(y_i), L_i)
\end{equation}
where $P(y_i)$ is the network output for input $x_i$, computed using the Softmax function, $L_i$ is the one-hot encoded vector for sample $i$, and $D(P(y_i), L_i)$ is the cross-entropy distance function defined as:
\begin{equation}
D(P(y_i), L_i) = - \sum_{j=1}^{K} l_{ij} \log(p_(y_{ij}))
\end{equation}
\par
The goal of the multinomial logistic classifier is to construct a model that learns the relationship, captured by the model weights, between the input features (i.e input RSS variables $x_i$) and the grid cells probability distribution $P(y_i)$, 
so that the outcome of a new RSS vector can be correctly predicted for a new data point for which the input RSS variables, but not the outcome, are available. 
Strictly speaking, the classifier attempts to learn the \textit{joint distribution} of the RSS coming from different cell towers. 
This distinguishes our deep model-based localization from the traditional probabilistic graphical models such as the Bayes network (e.g in \cite{ibrahim2012cellsense}) which learns the distribution of each cell-tower RSS assuming independence of the cell towers. 

\noindent \textbf{ Increasing Model Robustness and Noise Handling}

To further reduce the effect of the inherent noise in the input training samples and increase the model robustness, \sys{} employs a drop-out regularization technique during training \cite{srivastava2014dropout}. The idea is to randomly drop neurons from the network during training (shaded nodes in Figure~\ref{fig:structure}). The temporary removed neurons no longer contribute to the activation of downstream neurons in the forward pass. Similarly,  the weight update process is not applied to them in the backward pass. This prevents the network from over-fitting the training data. 

\subsection{The Online User Tracking Phase}
During the online user tracking phase, \sys{} \textit{depends only on the cellular RSS} coming from the RSS Collector module of a user standing at unknown location.

To estimate the user location, the Location Estimator first estimates the discrete grid cell $g^* \in \mathbb{G}$ in which the user exists given the received RSS vector $x=(x_1, x_2,...x_M)$. 
To this end, \sys{} leverages the trained deep model. Specifically, the input vector is fed through the deep network and 
the grid cell with the maximum probability is selected as the the estimated grid cell. More formally, 

\begin{equation}
g^* = \underset{g \in \mathbb{G}}{\operatorname{argmax}}[P(g|x)]
\end{equation}

The center of mass of $g^*$ represents a coarse-grained location estimate of the user location. To further refine the estimated position in the continuous space, the final user estimated location $l^*$ is estimated as the center of mass of all grid-cells, weighted by their respective probability:
\begin{equation}
l^* =\sum_{g \in \mathbb{G}} g \times p_g
\end{equation}
where $\mathbb{G}$ is the universe of grid cells in the area of interest and $p_g$ is the corresponding estimated probability at the output layer of the model. %
\section{Performance Evaluation}
\label{evaluation}
In this section, we evaluate \sys{} performance in both urban and rural environments. We start by studying the effect of different parameters on \sys{} performance. Then, we quantify the overall system performance and compare it to the state-of-the-art techniques in both rural and urban areas. 
\begin{figure*}[!t]
	\minipage{0.32\textwidth}
	\includegraphics[width=\linewidth]{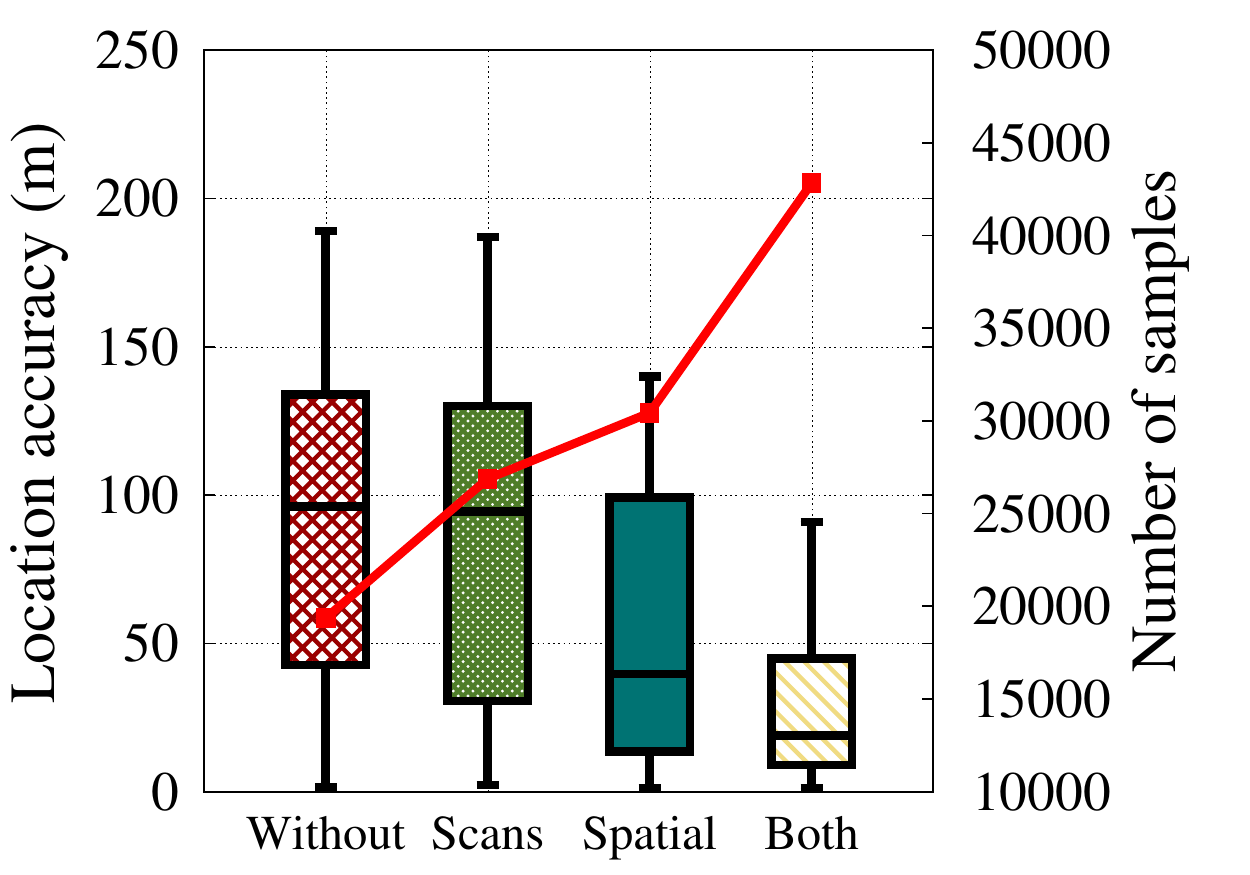}
	\caption{Effect of different data augmentation techniques on system performance.} %
	\label{fig:augmentation}
	\endminipage\hfill
	\minipage{0.32\textwidth}
	\includegraphics[width=\linewidth]{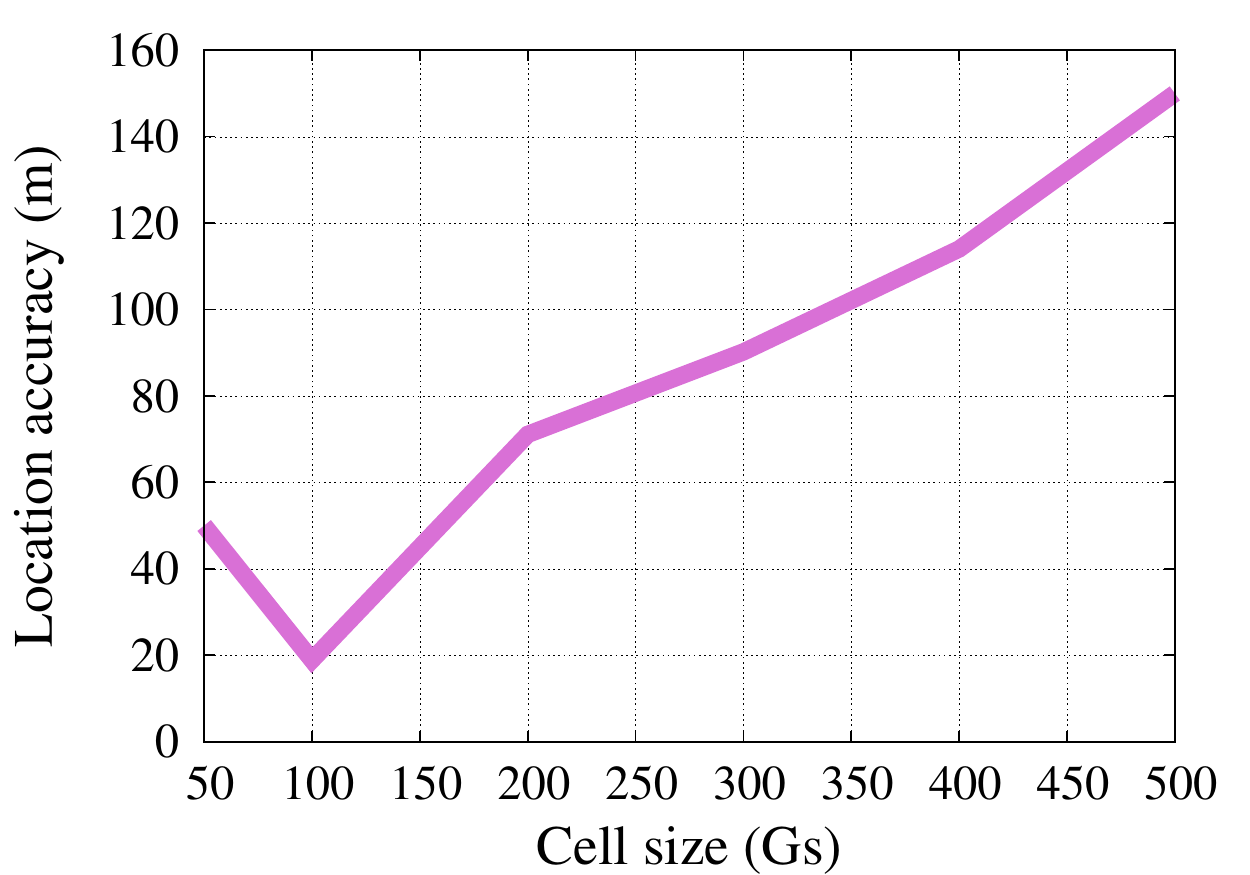}
	\caption{Effect of changing grid cell length on the localization accuracy.}
	\label{fig:cellsize}
	\endminipage\hfill
	\minipage{0.32\textwidth}
	\includegraphics[width=\linewidth]{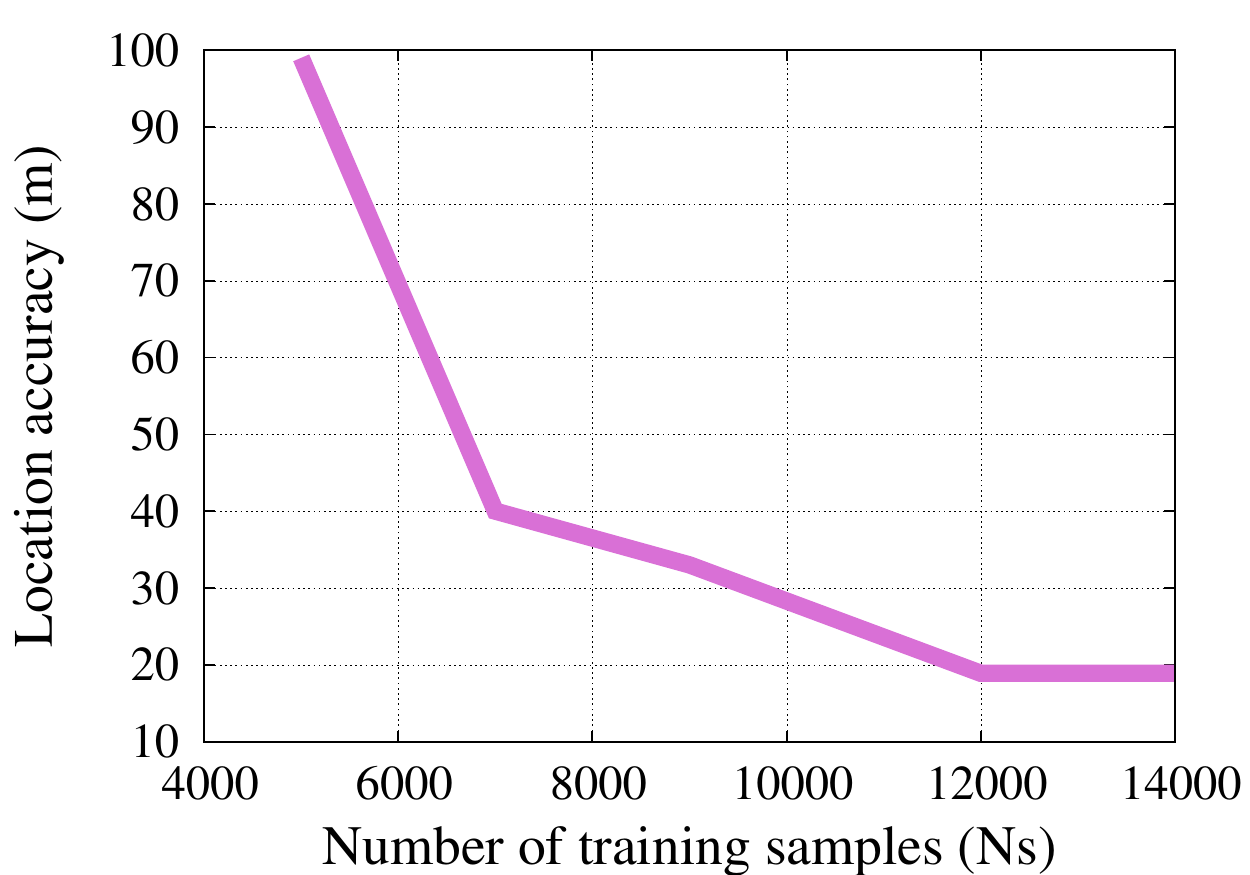}
	\caption{Effect of changing number of training samples on the localization accuracy.}
	\label{fig:trainingsamples}
	\endminipage\hfill
\end{figure*}
\subsection{Data collection}
We collected data for two different testbeds. The first covers a 0.2$\text{Km}^2$ urban area while the second covers a 1.2$\text{Km}^2$ rural area. 
Table~\ref{testbeds} summarizes the characteristics of both testbeds. 
\begin{table}[!t]
	\centering
	\caption{Testbeds}
	\label{testbeds}
	\begin{tabular}{|l||l|l|p{1.7cm}|}\hline
		\textbf{Testbed} &\textbf{ Area size} ($\text{Km}^2$) & \textbf{Trace samples} & \textbf{Cell towers density}\\ \hline \hline
		\textbf{Urban} & 0.2  & 19,369 & 185 \\ \hline
		\textbf{Rural} & 1.2  &  44,659 & 20 \\ \hline
	\end{tabular}
\end{table}

Data is collected by war-drivers using different Android devices including HTC Nexus One, Prestigio Multipad Wize 3037 3G, HTC One X9 and Motorola Moto G5 plus phones among others.
The data collectors' devices have GPS receivers, which we used for collecting the ground-truth locations. The deployed collector software collects GPS ground-truth locations, received signal strengths, and timestamps. The sampling rate is one sample per second.

For training the deep network model, We tried different network structures. The best structure contains three hidden layers with 40, 256, and 40 neurons. This structure produces a consistent accuracy for both urban and rural areas. We study the effect of the training epochs and learning rate on the system performance later in this section.

\subsection{Effect of \sys{} Parameters}
In this section, we study the effect of different parameters on \sys{} 
performance including the effect of the different data augmentation techniques, changing grid spacing, changing the number of samples used in training, cell towers density, number of training epochs, and learning rate. 

Table \ref{default} summarizes the system parameters used throughout the experiments and their default values.

\begin{table}[!t]
	\centering
	\caption{Parameters and their default values}
	\label{default}
	\scalebox{0.7}{
		\begin{tabular}{|l||l|p{2cm}|}\hline
			\textbf{Parameter} & \textbf{Range} & \textbf{Default value}\\ \hline \hline
			Data augmentation technique & \pbox{10cm}{Without augmentation,\\Spatial augmentation,\\Scans augmentation,\\ Spatial \& Scans augmentation}  & Spatial \& scans augmentation\\ \hline
			Grid cell length ($G_s$) & 50-500m  & 100m\\ \hline
			Number of training samples ($N_s$) & 5000-15000 & 19368 \\ \hline
			Number of training epochs ($N_e$) & 500-3000 & 3000 \\ \hline
			Learning rate ($\alpha$) & 0.01-0.001  & 0.005\\ \hline
			Cell towers density ($D_s$) &  25\%-100\% & 100\% \\ \hline
			Testbed & Urban and rural & Urban\\ \hline
		\end{tabular}
	}
\end{table}

\subsubsection{Effect of different data augmentation techniques}
Figure~\ref{fig:augmentation} shows the effect of different data augmentation  techniques described in Section~\ref{sec:data_aug} on the system accuracy. The figure shows that both scans and spatial augmentation can improve the system accuracy compared to training without augmentation. The figure also shows that the accuracy is correlated with the number of samples generated by the augmentation technique. Applying both augmentation techniques together leads to the best accuracy  with higher accuracy. This confirms that the spatial and scans augmentation can significantly improve the accuracy, not only by increasing the number of training samples, but also by implicitly handling the inherent noise in the input GPS locations and RSS values. %

\begin{figure*}[!t]
	\minipage{0.32\textwidth}
	\includegraphics[width=\linewidth]{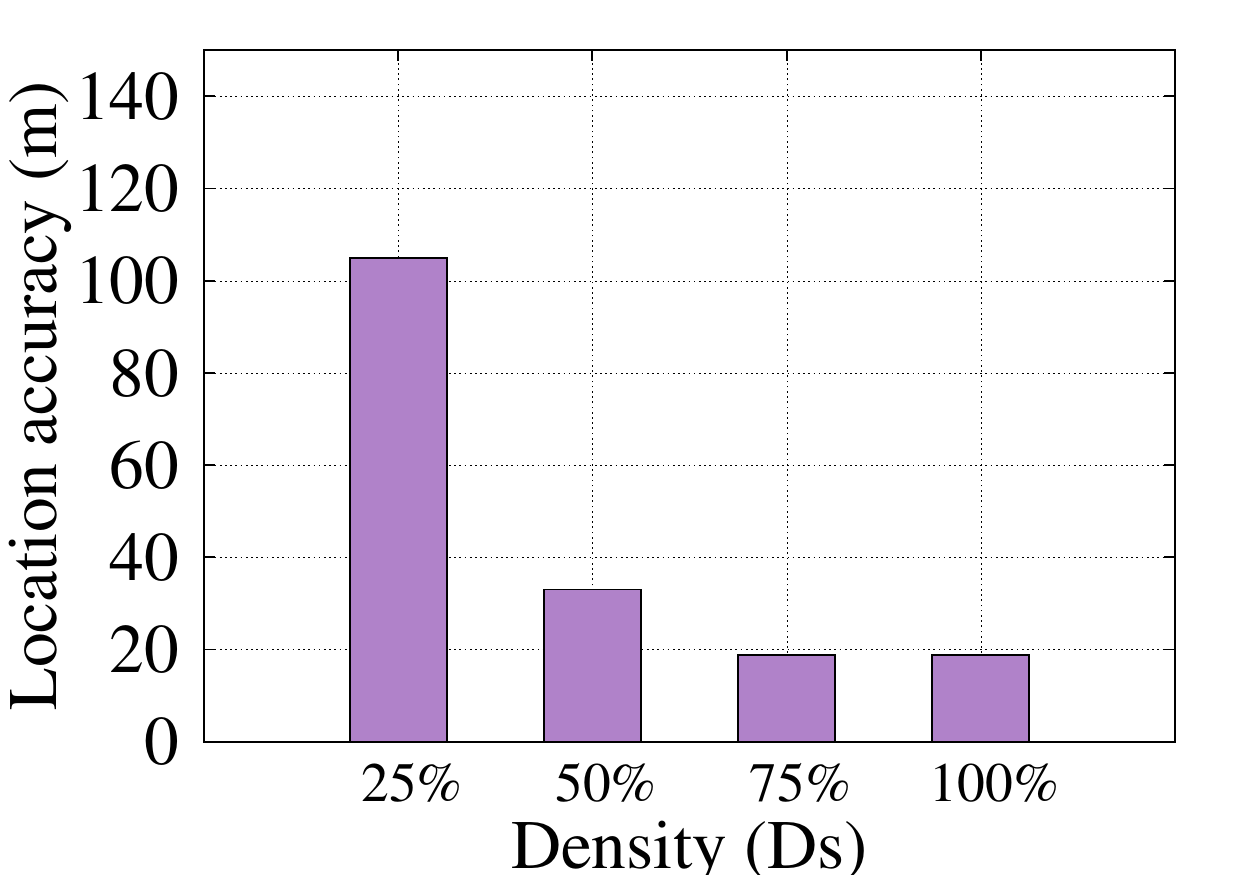}
	\caption{Effect of changing cell towers density on the localization accuracy.}
	\label{fig:density}
	\endminipage\hfill
	\minipage{0.32\textwidth}
	\includegraphics[width=\linewidth]{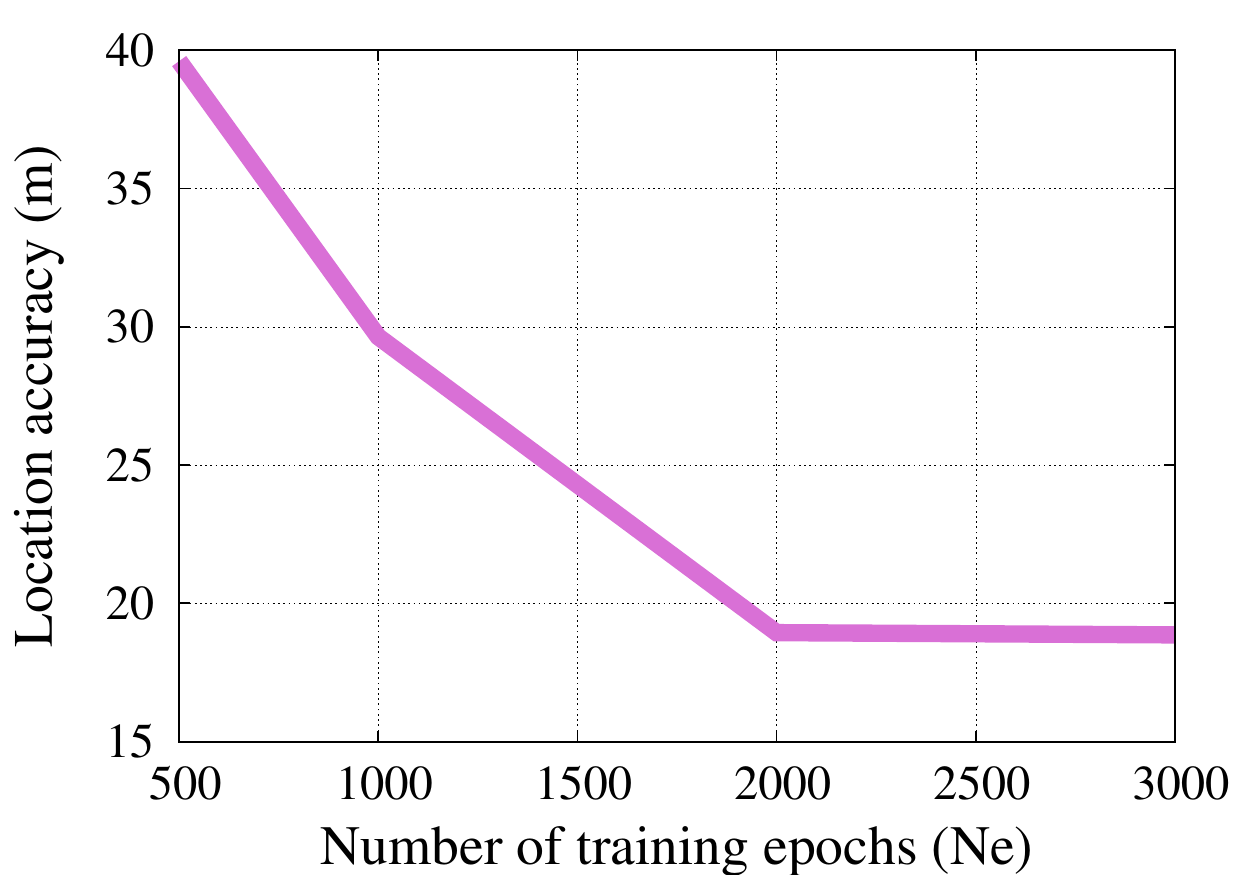}
	\caption{Effect of changing number of training epochs on the localization accuracy.}
	\label{fig:epochs}
	\endminipage\hfill
	\minipage{0.32\textwidth}
	\includegraphics[width=\linewidth]{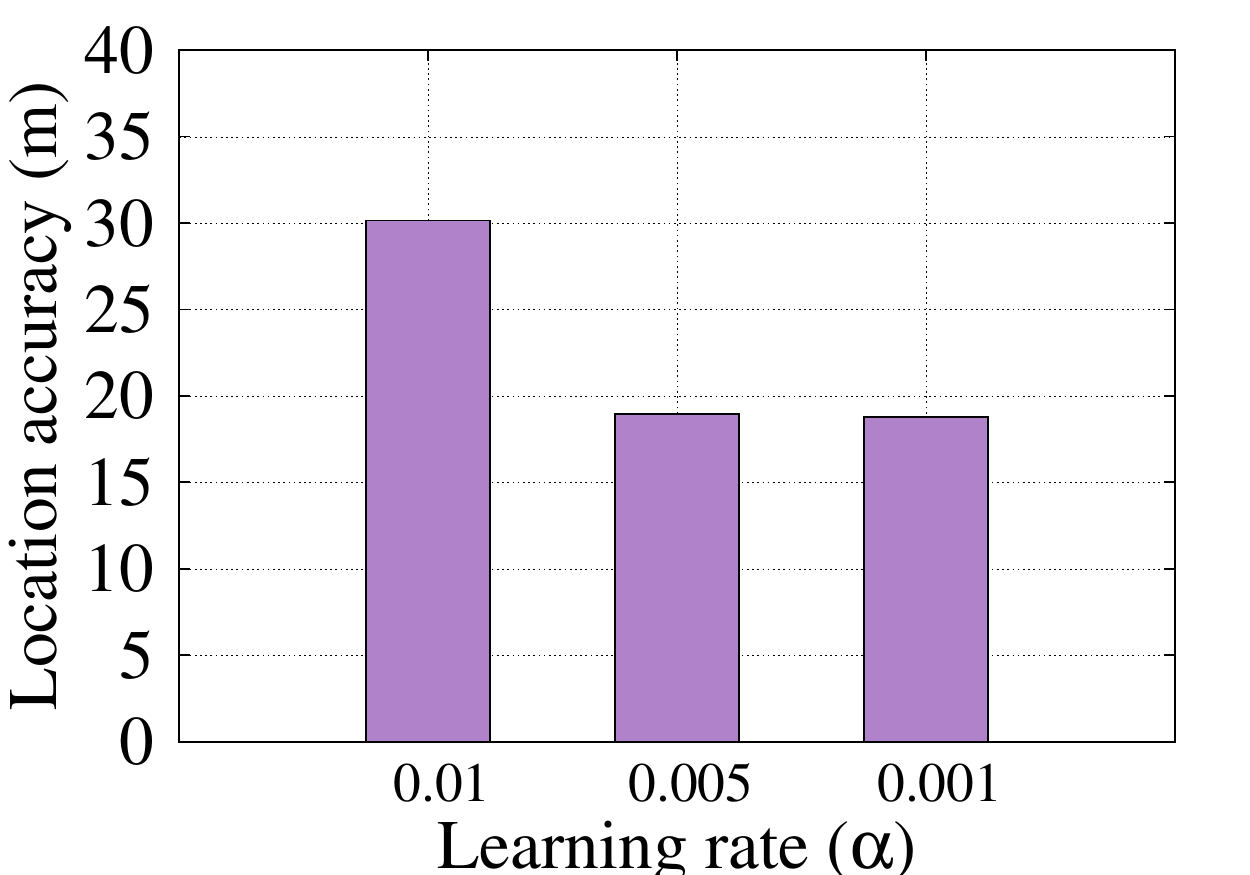}
	\caption{Effect of changing learning rate on the localization accuracy.}
	\label{fig:learningrate}
	\endminipage\hfill
\end{figure*}
\subsubsection{Effect of virtual grid spacing ($G_s$)}
Figure \ref{fig:cellsize} shows the effect of changing the grid cell length (parameter $G_s$) on the system median localization accuracy.
The figure shows that as the grid cell length increases, the system accuracy increases until it reaches an optimal value at $G_s=100m$ then the performance degrades with the increase of the grid cell length. This can be explained by noting that as the grid cell length increases, we have more training samples per cell, leading to better models. However, as the grid cell length increases the grid cells cover larger areas, leading to coarser-grained accuracy. This trade-off leads to the optimal value at $G_s=100m$.

\subsubsection{Effect of number of samples used in training ($N_s$)}
Figure~\ref{fig:trainingsamples} shows the effect of increasing the number of  training samples on the system accuracy. The figure shows that, as expected, increasing number of training samples increases the system accuracy due to the increase of the quality of the learned model. The good news is that the system performance saturates after about 12000 samples. This can be further reduced using the data augmentation module as quantified before.  

\subsubsection{Effect of cell towers density ($D_s$)}
Figure~\ref{fig:density} shows the effect of cell towers density percentage on the system accuracy. To reduce the cell tower density, we remove a percentage of the cell towers as indicated in the figure. The figure shows that increasing the cell towers density leads to better accuracy due to the addition of new features (the RSS from the new towers). These features lead to better  discrimination between the different grid cells, which in turn results in a better accuracy. \sys{} maintains its accuracy even with 75\% of the typical density.

\subsubsection{Effect of number of training epochs ($N_e$)}
Figure~\ref{fig:epochs} shows the effect of changing the number of training epochs on the system accuracy. This parameter controls the number of times the system iterates over the entire training dataset to update the weights.
The figure shows that increasing number of training epochs leads to a better localization accuracy. Nevertheless, increasing number of training epochs over 2000 has no effect on the system accuracy.

\subsubsection{Effect of changing learning rate ($\alpha$)}
Figure~\ref{fig:learningrate} shows the effect of changing the learning rate on the localization accuracy. This parameter enables the deep learning training optimizer to control the weights update in the direction of the gradient during the optimization progress.
The figure shows that lowering the learning rate to ($\alpha = 0.005$) %
makes the training more reliable. This is because lowering the learning rate makes the optimizer step towards the minimum slowly which prevents overshooting it. Note that the network training is performed during the offline phase, which allows us to increase the number of epochs for better training without overhead. 
\subsection{Overall System Performance}
In this section, we study the effect of deploying \sys{} with the default parameters
shown in Table~\ref{default}. %

\subsubsection{\sys{} accuracy in urban and rural areas}
\begin{figure}[!t]
	\includegraphics[width=0.7\linewidth]{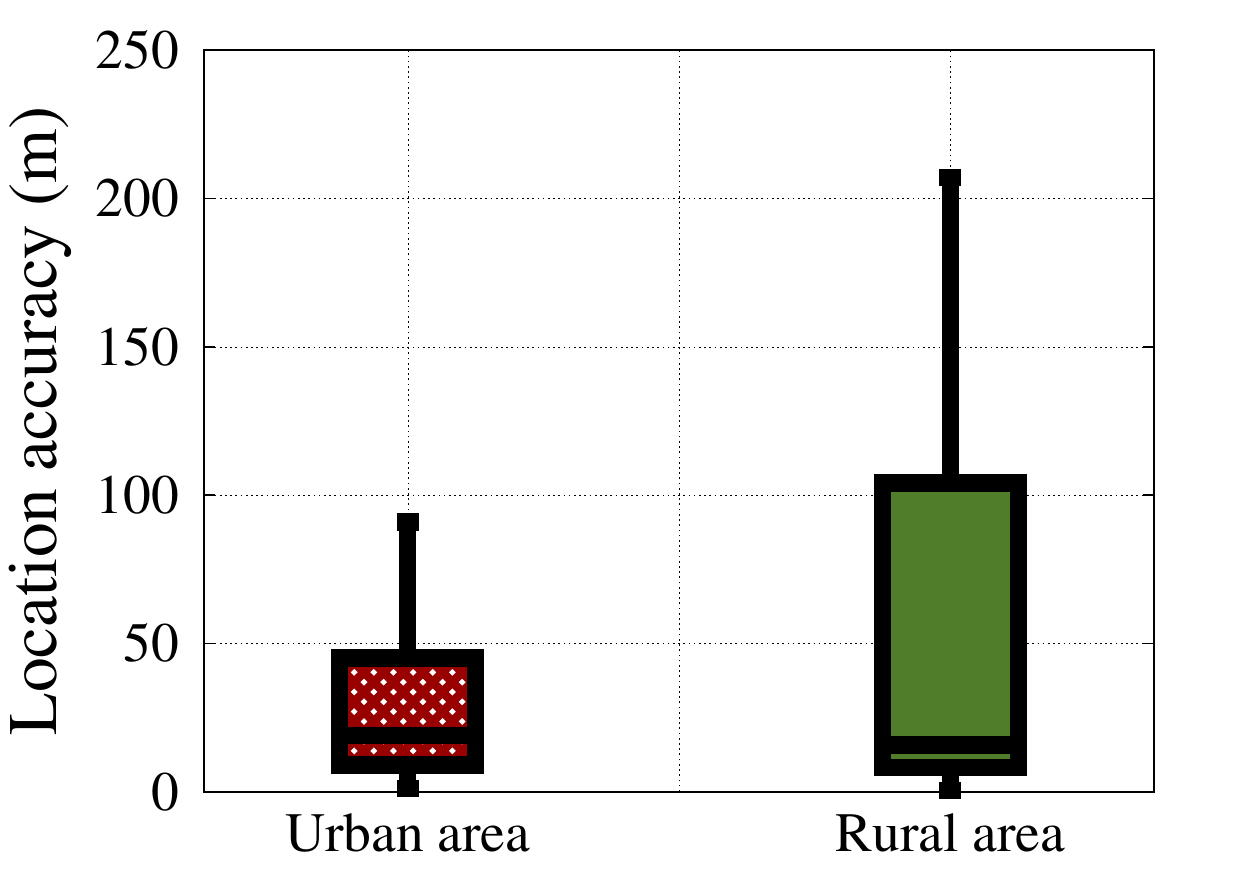}
	\caption{A comparison between the system accuracy in urban and rural areas.}
	\label{fig:urbrur}
\end{figure}

Figure~\ref{fig:urbrur} shows the box plot for \sys{} localization error in urban and rural areas. The figure shows that the median accuracy of both testbed is consistent: 18.8m in urban and 15.74m in rural. The variance of the localization error in the rural area is higher with a longer distribution tail. This is expected due to the lower density of cell towers in the rural dataset as compared to the urban dataset.  
\subsubsection{Comparison with other systems}
In this section we compare \sys{} performance to the state-of-the-art  probabilistic graphical model-based technique (i.e the \textit{CellSense} system \cite{ibrahim2012cellsense}) that provides the best performance of the current outdoor cellular localization systems. \textit{CellSense} uses a probabilistic fingerprint for localization. In particular, assuming the independence of RSS from different cell towers, it stores the RSS histogram for each cell tower in the area of interest at each grid cell. These histograms are then used during the online tracking phase to estimate the most probable cell location.

\begin{figure*}[!t]
	\centering
	\begin{subfigure}[b]{0.4\textwidth}
		\includegraphics[width=1\textwidth]{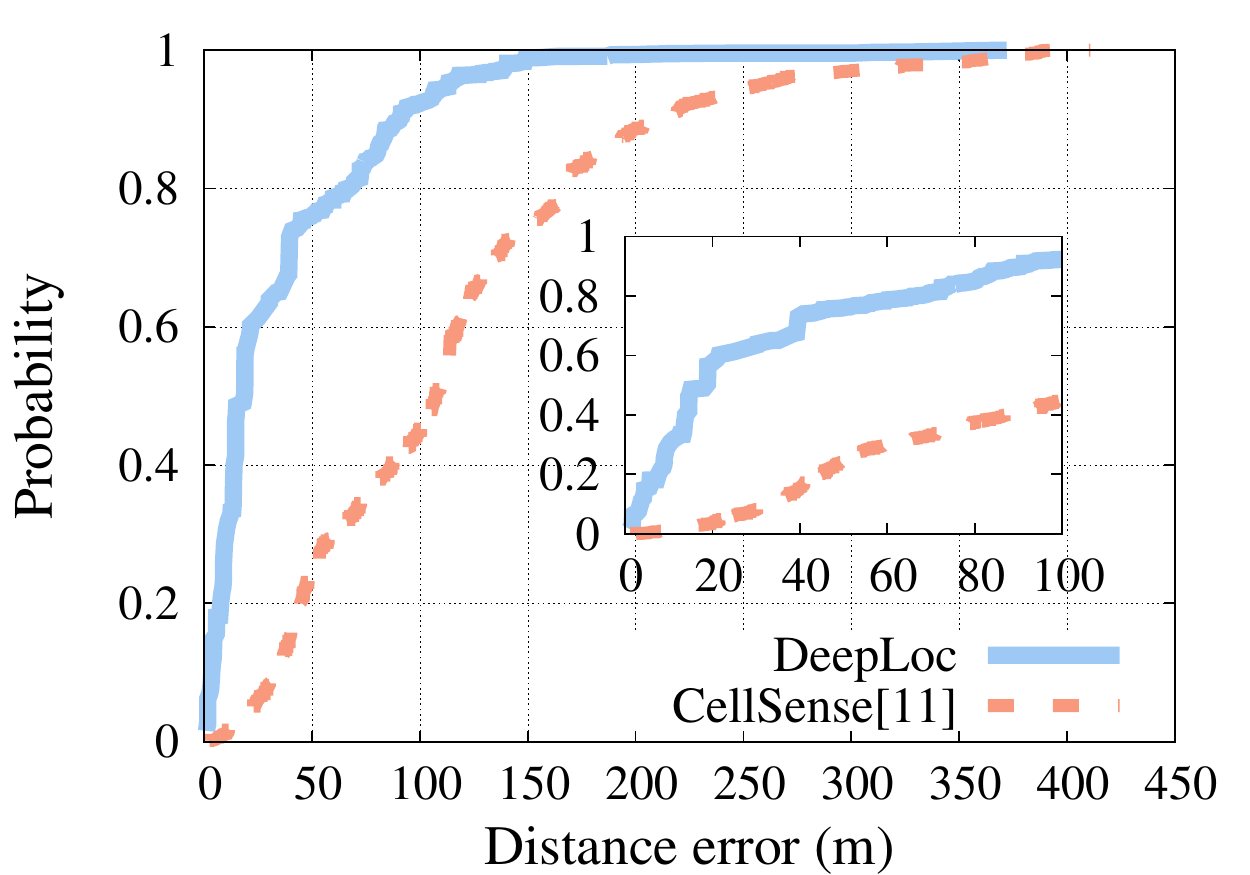}%
		\caption{Urban area.}
		\label{fig:cdfs-urb}
	\end{subfigure}%
	\begin{subfigure}[b]{0.4\textwidth}
		\includegraphics[width=\textwidth]{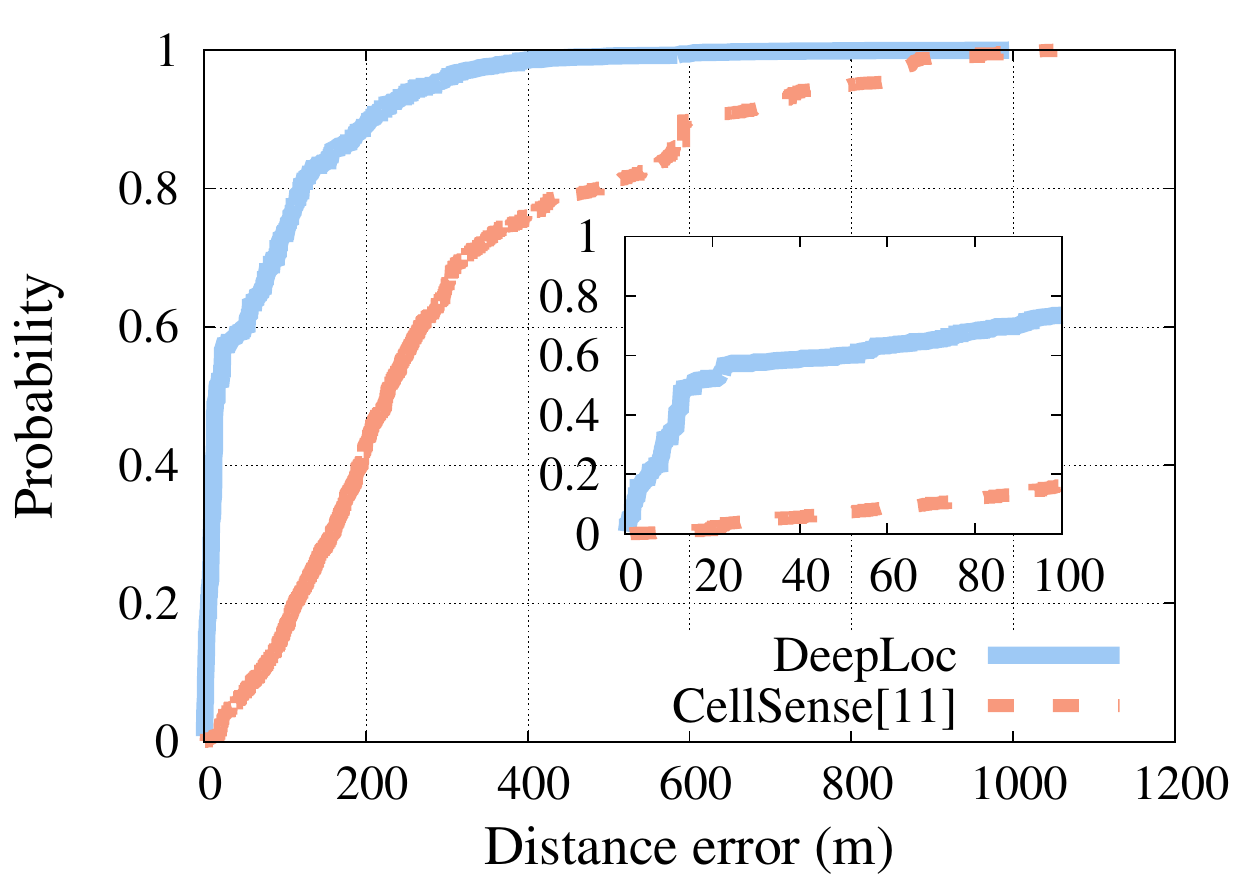}
		\caption{Rural area.}
		\label{fig:cdfs-rur}
	\end{subfigure}
	\caption{CDF of localization error.}
	\label{fig:cdfs}
\end{figure*}

Figure~\ref{fig:cdfs} shows the CDF of distance error for both systems in urban and rural areas. Table~\ref{comparison} summarizes the results. The figure shows that our deep learning based system, \sys{}, 
can achieve significantly better accuracy than the traditional graphical model-based system  by more than 471\% and by 1337\% in urban and rural areas, respectively. Moreover, \sys{} enhances all the other quantiles. This confirms that the deep learning model can capture the interrelated signatures of the different cell towers at different grid cells better than probabilistic models that assume independence. 

\subsubsection{Power Consumption}
\begin{figure}[!t]
	\includegraphics[width=0.95\linewidth]{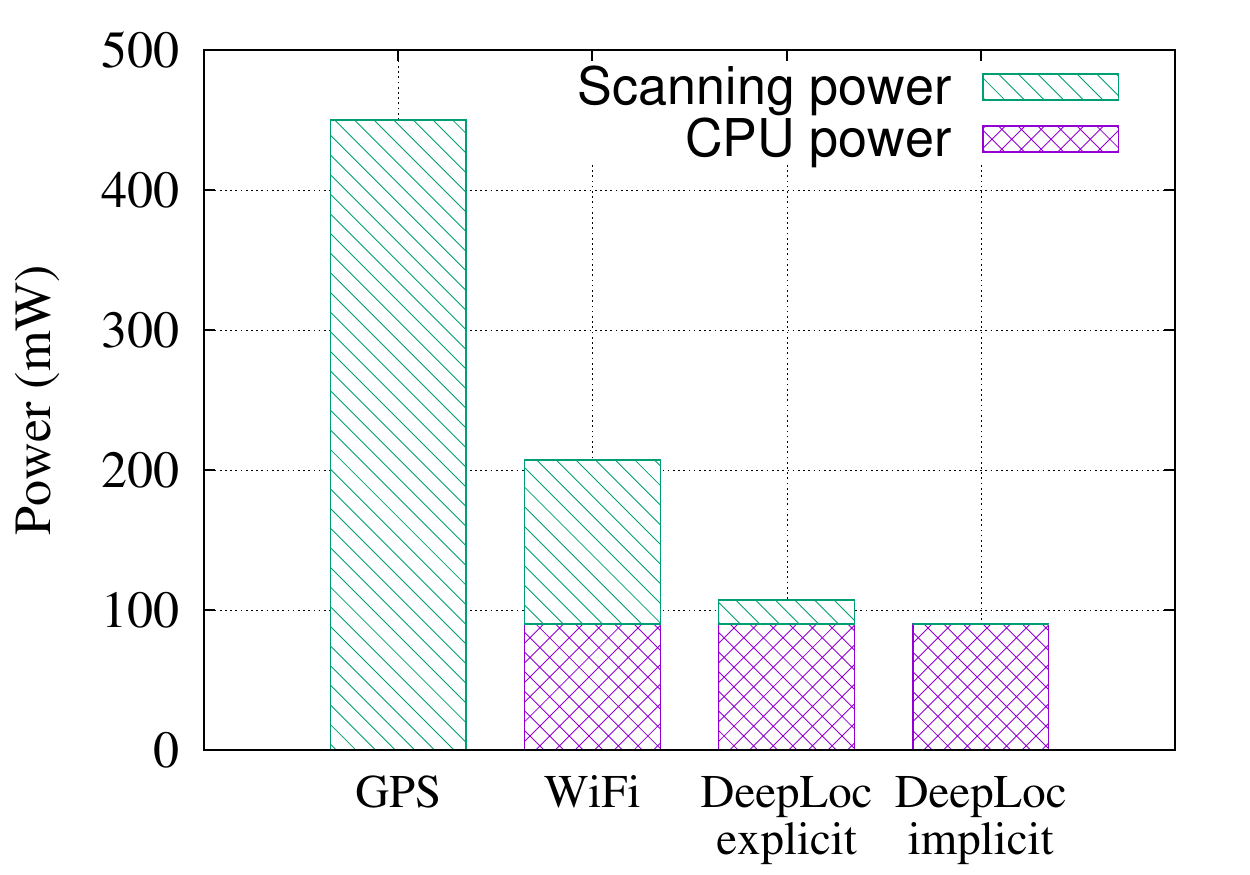}
	\caption{Power consumption of the different systems.}
	\label{fig:power}
\end{figure}

Figure~\ref{fig:power} shows the CPU and scanning power consumption of \sys{} compared to using GPS or WiFi. The figure shows that \sys{} power consumption is significantly less than that of using GPS or WiFi by at least 50\%. In addition, since \sys{} depends on the cellular signal, which is a main functionality of any cell phone; it consumes virtually zero extra energy compared to the standard phone operation. 

\begin{table*}[!t]
	\centering
	\caption{Comparison between CellSense and DeepLoc}
	\label{comparison}
	\begin{tabular}{|l|l|l|l|l|l|}\hline
		Accuracy & Minimum &\pbox{10cm}{$25^{th}$\\percentile }& \pbox{10cm}{$50^{th}$\\percentile}&\pbox{10cm}{$75^{th}$\\percentile}& Maximum\\ \hline \hline
		\textbf{DeepLoc (urban)} & \textbf{1.1} & \textbf{8.97}  & \textbf{18.8} & \textbf{44.98} & \textbf{372}\\ \hline
		CellSense\cite{ibrahim2012cellsense} (urban) & 2.38 (-116\%) & 50.4 (-462\%) &  107.4 (-471\%) & 151.6 (-237\%) & 410 (-10\%)\\ \hline \hline
		\textbf{DeepLoc (rural)} & \textbf{0.34} & \textbf{8.1}  & \textbf{15.74} & \textbf{104} & \textbf{995}\\ \hline
		CellSense\cite{ibrahim2012cellsense} (rural) & 1.1 (-223\%)  & 135 (-1567\%) &  225.58 (-1337\%) & 385 (-270\%) & 1050 (-5.5\%)\\ \hline \hline
	\end{tabular}
\end{table*}
\vspace*{-.2cm}

\section{Related Work}
\label{history}

\sys{} leverages RSS samples from the different cell towers to train a deep neural network model to achieve accurate outdoor localization. In this section, we discuss the localization techniques literature and how they differ from \sys{}. 

\subsection{GPS-based Techniques}
The Global Positioning System (GPS) \cite{hofmann2012global} is the most commonly used outdoor localization technique. Although it can provide outdoor location worldwide, it suffers from a number of drawbacks. First, it consumes a lot of power. Second, it requires a direct line-of-sight with th satellite, limiting its coverage in areas; e.g. with tall buildings and inside tunnels; and even in bad weather conditions \cite{cui2003autonomous,aly2013dejavu,aly2017dejavuj}. To reduce the high energy consumption of GPS, a number of techniques have been proposed that combine other techniques, e.g. inertial sensors or map matching \cite{aly2015semmatch,mohamed2014accurate}, with a duty cycled GPS \cite{youssef2010gac,constandache2009enloc}. However, those techniques trade the accuracy with low power consumption. In addition, GPS and inertial sensors are still not available in a variety of low-end phones deployed throughout the world, limiting their ubiquitous availability.

\textit{DeepLoc, on the other hand, is an energy-efficient technique that can provide a highly accurate localization in different environments. In addition, cellular RSS is available in all cell phones, providing ubiquitous coverage.}

\subsection{Sensors-based Techniques}
Another set of techniques leverage WiFi APs \cite{cheng2005accuracy,lamarca2005place,Skyhook,rekimoto2006placeengine,yoshida2005locky} or augmented sensors on the smart-phones \cite{aly2015lanequest,aly2013dejavu,azizyan2009surroundsense,constandache2010towards,ofstad2008aampl,abdelnasser2016semanticslam,wang2012no} to localize users, without using the GPS. The idea is to use a WiFi fingerprint or landmarks detected by the different phone sensors to localize the phone. These systems can provide remarkable accuracy with a low-energy profile. However, WiFi-based techniques suffer from coverage problems as WiFi signals are weak outdoors, limiting their ability to differentiate between distinct locations. In addition, there are many locations that are not covered by WiFi, such as on highways and rural areas. %
 Furthermore, both WiFi and augmented sensors-based techniques are not ubiquitously available on all cell phones. %

\textit{\sys{}, on the other hand, inherits the ubiquitous coverage of cellular networks and is available on all cell phones.} %
	
	\subsection{Cellular-based Techniques}
	Cellular-based localization systems have been proposed as an alternative to provide ubiquitous localization. They are either Cell-ID based \cite{paek2011energy} or RSS-based \cite{ibrahim2012cellsense,ibrahim2011hidden,constandache2009enloc,ibrahim2010cellsense,laoudias2018survey}. Cell-ID techniques estimate the location of the user as the location of the strongest cell tower heard by the user's device. They are simple to deploy but provide coarse-grained accuracy. RSS-based techniques leverage the cellular RSS to localize the user. To handle the noise in th wireless signal, these techniques employ probabilistic graphical models to infer the user's position, e.g. Baysian-based models as in \cite{ibrahim2012cellsense,ibrahim2011hidden,ibrahim2010cellsense}. These probabilistic models should optimally capture the joint distribution of the different cell towers in the area of interest at different fingerprint locations. However, to make the problem mathematically tractable and avoid the curse of dimensionality \cite{bishop2006pattern}, these systems usually assume that the different cell towers are \textit{independent}, limiting their ability to capture the inherently correlated relation between the RSS from the different cell towers, which in turn lowers their accuracy.
	
	\textit{DeepLoc, on the contrary, leverages a deep-model based technique to  learn the joint distribution of input features, which leads to significantly better performance than traditional graphical-based models. In addition, it has provisions to handle the noise in the input GPS locations and RSS.}
	
	\subsection{Deep Models-based Techniques}
	Recently, a number of deep learning based fingerprinting techniques have been proposed \cite{felix2016fingerprinting,laoudias2009localization,wang2015deepfi,wang2017csi,Ibrahim2018} in \textit{indoor environments}. These systems  collect the fingerprint/training samples manually, using stationary receivers that need to stand for a few minutes at each discrete fingerprint locations. This high overhead process does not scale to the large areas characterizing outdoor environments. In addition, none of the previous systems address the large amount of data required for training a deep learning model. %
	
	\textit{\sys{}, on the other hand, addresses the scalability issues of outdoor environments through the virtual griding approach. In addition, it introduces spatial and scan data augmentation techniques to reduce the number of samples required to train a deep learning model. Moreover, its design includes a number of provisions to handle the noise in the input data as well as provide robust performance in different environments.} 
	
	\section{Conclusion}
	\label{conclusion}
	We presented \sys{}: a deep learning-based system for outdoor localization. \sys{} has different techniques that can augment the training data to reduce the calibration overhead as well as handle the noise in the GPS and RSS data. We showed how the griding-based approach can further remove the requirement of the data collectors to stand at specific locations in the area of interest. Finally, we showed how \sys{} includes provisions in the model to avoid over-training and increase the model robustness.
	\par
	Implementation of \sys{} on different Android phones in urban and rural areas showed that \sys{} can achieve a consistent median localization accuracy of 18.8m in urban areas and 15.7m in rural areas. This accuracy is comparable to GPS accuracy and significantly better than the traditional graphical models-based localization systems by more than 470\% in different environments. Moreover, \sys{} power consumption is less than GPS by 330\%.
	\vspace*{.1cm}
	
	\begin{acks}
	This work is supported in part by a grant from the Egyptian National Telecommunications Regulatory Authority (NTRA).
	\end{acks}

\bibliographystyle{ACM-Reference-Format}
\bibliography{main.bbl}

\end{document}